\documentclass{article}

\usepackage[preprint]{neurips_2026}
\usepackage{minitoc}
\usepackage{neurips_2026}
\usepackage{amsmath}
\usepackage{graphicx}
\usepackage{enumitem}
\usepackage{subcaption}
\usepackage{wrapfig}
\usepackage{algorithm}
\usepackage{algpseudocode}  

\usepackage[utf8]{inputenc} 
\usepackage[T1]{fontenc}    
\usepackage{hyperref}       
\usepackage{url}            
\usepackage{booktabs}       
\usepackage{amsfonts}       
\usepackage{nicefrac}       
\usepackage{microtype}      
\usepackage{xcolor}         
\usepackage{float}
\usepackage{stfloats}
\usepackage{amsthm}
\usepackage{graphicx}
\usepackage{wrapfig}
\usepackage{bbm}
\usepackage{caption}
\usepackage{subcaption}
\usepackage{ragged2e} 
\usepackage{enumitem}
\usepackage{balance}
\usepackage{booktabs,makecell, multirow, tabularx}
\usepackage[font=small,labelfont=bf]{caption}  

\newcommand{\ie}{\emph{i.e., }}
\newcommand{\eg}{\emph{e.g., }}

\title{Kara: Sliding-Window KV Cache Compression for Efficient Serving of Reasoning LLMs under Memory Constraints}

\author{%
  Han Shen\thanks{Han Shen and Yuyang Wu are co–first authors and contribute equally to this paper.} \\
  \texttt{kingsutherszsz48@gmail.com}
  \And
  Yuyang Wu\footnotemark[1] \\   
  Carnegie Mellon University \\
  \texttt{yuyangwu@cs.cmu.edu} \\
  \And
  Junpu Yu\\
  Carnegie Mellon University\\
  \And
  Olexandr Isayev\thanks{Corresponding Author}\\
  Carnegie Mellon University
}

\begin{document}

\maketitle

\begin{abstract}
Reasoning language models often generate long chain-of-thought (CoT), which accumulates a massive KV cache during the decoding phase and incurs high decoding latency and limited output throughput.
To address these issues, KV cache compression has emerged as a promising technique for reducing memory footprint by selectively removing unimportant KV pairs while preserving useful ones for subsequent decoding.
Nevertheless, we identify two key limitations in existing KV cache compression methods: 1) Their threshold-triggered compression policy may reduce output throughput under memory-constrained concurrent serving, and may fully eliminate KV pairs from certain blocks of the sequence, potentially worsening information loss.
2) They typically retain either isolated KV pairs or fixed-size chunks with rigid boundaries, failing to preserve flexible-sized chunks at arbitrary token positions.
To overcome these limitations, we propose Kara, a sliding-window KV cache compression method that performs decoding-time compression by operating only on the recently generated context.
Kara leverages bidirectional attention to score and select informative KV pairs in the window. 
To enable flexible preservation of important semantic information, we design a Token2Chunk module to expand a subset of selected KV pairs into chunks.
Furthermore, we adapt Kara to PagedAttention and develop KvLLM, an inference framework built upon vLLM, which reduces KV cache memory usage and improves output throughput in memory-constrained environments.
Kara preserves nearly 100\% of the full-KV accuracy while retaining only 20\% of the KV cache, and KvLLM improves throughput by 12.75\% on average over vanilla vLLM under memory-constrained concurrent serving.
\end{abstract}

\section{Introduction}
Chain-of-thought reasoning \cite{guo2025deepseek,wei2022chain,yang2025qwen3,banerjee2025crane} has endowed LLMs with powerful capabilities for solving complex tasks. 
However, reasoning models often generate long traces, which lead to a substantial growth of KV cache during decoding and impose significant memory overhead \cite{li2024survey,yuan2024llm}. 
For example, caching 128K tokens in Qwen3-14B with FP16 precision requires approximately 20 GB of memory \cite{yang2025qwen3}. 
This overhead becomes especially pronounced in memory-constrained concurrent serving settings, where the rapidly growing KV cache increases decoding latency and causes input requests to wait for available memory resources, thereby reducing overall output throughput.

To address these issues, KV cache compression~\cite{cai2026rkv,feng2026adakv,li2024snapkv}, an inference-time approach, has garnered increasing attention. 
By selectively retaining important KV pairs, KV cache compression can reduce memory usage while keeping performance degradation within an acceptable range (\eg within a 1\% performance drop) \cite{yuan-etal-2024-kv}.
Existing works \cite{li2024snapkv,yuan-etal-2024-kv} have shown strong performance on multiple tasks while retaining only a small fraction of the KV cache (\eg 30$\%$). 
Early compression methods \cite{xiao2024efficient,zhang2025beyond} relied on heuristic techniques such as attention sinks to preserve KV pairs at fixed positions, ignoring the varying importance of different tokens and resulting in suboptimal performance. 
More recently, research has shifted toward score-based KV cache compression \cite{cai2026rkv,li2024snapkv,dms,kim2026kvzip}, which assigns importance scores to multi-head KV pairs to determine cache retention. 
These methods typically maintain a query cache of recent query states and use them to score historical KV pairs via attention with their key states. 
The score-based compression methods often achieve state-of-the-art performance, as they leverage the internal information of LLMs to identify critical KV pairs \cite{devoto2025expectedattention}.
\begin{figure}[t]
  \centering
  \begin{minipage}[t]{0.55\textwidth}   
    \centering
    \setlength{\abovecaptionskip}{-10pt}
    \setlength{\belowcaptionskip}{-15pt}

    \includegraphics[width=1.1\linewidth, trim=30 20 -30 0]{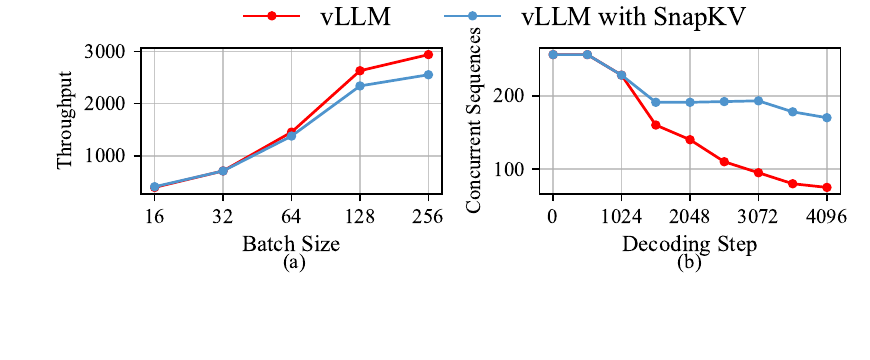}
    \caption{(a) Average output throughput under different batch sizes (\ie  the predefined maximum number of decoding sequences). 
    (b) The actual number of decoding sequences varies with decoding steps. 
    The decoding step denotes the number of global decoding iterations.
    The predefined maximum number of concurrently decoding sequences is set to 256.
    SnapKV constrains the KV cache size, freeing memory to decode more sequences concurrently.
    }
    \label{fig:F1}
  \end{minipage}
  \hfill
  \begin{minipage}[t]{0.40\textwidth}   
    \centering
    \setlength{\abovecaptionskip}{-10pt}
    \setlength{\belowcaptionskip}{-15pt}
\includegraphics[width=0.93\linewidth, trim=20 0 -10 40]{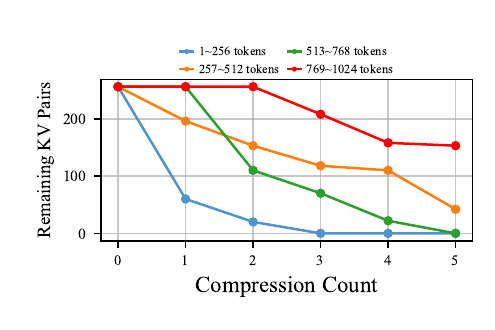}
    \caption{Remaining KV pairs under different compression counts. We track the remaining KV pairs of tokens from different regions for a certain attention head across repeated compression steps.
    We observe that the semantic information of certain regions is completely eliminated as compression proceeds.}
    \label{fig:F2}
  \end{minipage}
\end{figure}

Despite their success, we identify two key limitations in existing decoding-time KV cache compression methods:
\begin{itemize} [leftmargin=1.0em]
    \item \textbf{These methods mainly rely on a threshold-triggered compression policy, which can degrade both reasoning performance and inference efficiency.} Under the threshold-triggered compression policy, the KV cache is compressed back to a smaller target length once it reaches a predefined threshold, after which this grow-and-compress pattern repeats.
    While effective in low-frequency or personal deployment settings, this policy can cause a concurrency–throughput inversion effect under memory-constrained concurrent serving. 
    Specifically, when the number of concurrently decoding sequences approaches the gap between the threshold and the compressed cache length, compression may be triggered frequently, and the associated computational overhead can outweigh the throughput benefits even as the number of concurrently decoding sequences increases.
    In addition, since compression is applied over the entire KV cache, previously compressed regions of the sequence may be compressed again, which can further worsen information loss. 
    To empirically validate these challenges, we conduct a simple experiment using the vLLM\footnote{We use nano-vLLM for our experiments. 
    It retains the core mechanisms of vLLM, including PagedAttention, continuous batching, tensor parallelism, and recomputation, and is convenient for further development} framework \cite{kwon2023efficient} with a representative decoding-time compression method (SnapKV\cite{li2024snapkv}) on the MATH-500 dataset \cite{lin2024criticbench} with distilled Llama-8B models \cite{grattafiori2024llama}.
    As shown in Figure \ref{fig:F1}, vLLM with SnapKV achieves lower output throughput than vanilla vLLM as the batch size increases (\eg  an average 11.15\% drop at batch size 128), even though it consistently maintains a higher actual number of concurrently decoding sequences than vanilla vLLM.
    In Figure \ref{fig:F2}, the corresponding KV pairs of the earlier tokens are almost completely removed as compression is applied repeatedly.
    Detailed experimental settings are provided in Appendix \ref{appx:vllm_set}.
    \item \textbf{These methods retain either isolated KV pairs or fixed-size chunks with rigid boundaries, failing to preserve important flexible-sized chunks at arbitrary token positions.} Mainstream KV cache compression methods select isolated KV pairs for retention, which can lead to significant loss of semantic information \cite{liu2026chunkkv}.
    To address this issue, several methods \cite{liu2026chunkkv,blocksparse} further propose retaining contiguous KV pairs to form chunks. 
    For example, ChunkKV \cite{liu2026chunkkv} partitions the historical KV cache into chunks of the same length with fixed boundaries.
    However, such retention paradigms are often rigid and fail to capture flexible-sized KV information distributed at arbitrary positions across the sequence.
\end{itemize}

To overcome these limitations, we propose {Kara}, a sliding-window KV cache compression method that performs decoding-time compression only on the recently generated context.
Through empirical analysis, we find that the accumulated bidirectional attention received by each KV pair serves as an effective indicator of its importance and informativeness. 
Building on this observation, Kara uses the accumulated bidirectional attention score as the importance measure to identify discrete candidate KV pairs within the window.
To preserve flexible-sized contiguous KV pairs at arbitrary positions, we introduce a simple yet effective {Token2Chunk} module.
Given the candidate discrete KV pairs, Token2Chunk treats every two consecutive candidate KV pairs as the endpoints of a chunk, and filters chunks according to their length and the importance scores of the endpoint KV pairs. 
Finally, we preserve the candidate discrete KV pairs as well as all KV pairs inside the selected chunks.
In summary, Kara produces a compressed KV cache that combines both isolated KV pairs and chunk-level KV pairs within the sliding window.
Furthermore, we adapt Kara to PagedAttention \cite{kwon2023efficient} and develop {KvLLM}\footnote{KvLLM is built upon the nano‑vLLM framework.}, a vLLM-based \cite{kwon2023efficient} inference framework equipped with a periodic compression policy.
To avoid frequent compression triggering during decoding, the periodic compression policy compresses the trailing blocks of selected sequences every fixed number of decoding steps, which controls when compression is triggered.
Kara preserves nearly 100\% of the full-KV accuracy while retaining only 20\% of the KV cache, and KvLLM improves output throughput by 12.75\% on average over vanilla vLLM under memory-constrained concurrent serving.

{In summary, our main contributions are as follows:}
\begin{itemize}[leftmargin=1.0em]
\item We propose {Kara}, a KV cache compression method that uses sliding-window bidirectional attention to identify candidate discrete KV pairs and employs a {Token2Chunk} module to generate chunks with contiguous KV pairs.
\item We adapt the proposed method to PagedAttention and design the {KvLLM} inference framework with a periodic compression policy, which can effectively improve output throughput and  concurrency in memory-constrained environments compared with existing compression methods.
\item For Kara, we conduct reasoning and NIAH evaluations, showing superior accuracy under the same budget.
For KvLLM, we measure output throughput and latency, demonstrating improved efficiency over vLLM with the baseline compression method in memory-constrained settings.
\end{itemize}

\section{Related Work}
\subsection{KV Cache Compression}
Early compression methods, such as {StreamingLLM} \cite{xiao2024efficient}, preserve KV pairs at fixed prefix positions, ignoring the varying importance of different tokens.
Mainstream KV cache compression methods
\cite{cai2026rkv,li2024snapkv,ramachandran2026thinkv,jones2026quoka,mao2026icecache} leverage LLM-internal query signals to score historical KV pairs and typically achieve state-of-the-art compression performance.
Nevertheless, most methods rely on a {threshold-triggered} policy for decoding-time compression, which may degrade output throughput under memory-constrained concurrent serving.
Some works \cite{dms,trimkv} mitigate the issues induced by this policy by adopting delayed compression.
For example, DMS \cite{dms} scores KV pairs within a recent sliding window and executes eviction after tokens leave the window.
TRIM-KV \cite{trimkv} further introduces a time-decay mechanism on KV importance to avoid immediate eviction after leaving the window.
However, these methods may still require frequent eviction operations under concurrency and may eliminate all KV pairs from contiguous spans.
Although a few methods \cite{liu2026chunkkv,blocksparse} propose retaining fixed-length chunks with rigid boundaries, they ignore flexible-sized chunks at arbitrary positions.
Different from the above works, we score KV pairs using sliding-window bidirectional attention and introduce {Token2Chunk} to expand discrete KV pairs into flexible-sized chunks.
Note that the sliding window in some works \cite{park2026keydiff,liu2026chunkkv} is mainly used to protect recent tokens from compression, whereas in our method it specifies the tokens to be compressed.
Furthermore, we adopt a periodic compression policy to compress the trailing window of selected sequences every fixed number of decoding steps, controlling when compression is triggered.
\subsection{Sparse Attention}
Sparse attention \cite{zhang2025spargeattention,lu2026moba,lin2026twilight,xattn} aims to reduce the quadratic attention cost by selectively computing only a subset of attention interactions while omitting the others.
Previous works \cite{sun2025efficient} train models to produce a dynamic attention mask before each attention operation, where the mask indicates the currently useful KV information for attention computation and skips the rest. 
However, these sparse attention methods typically do not reduce the KV cache memory footprint, and they need to be applied at every decoding step \cite{sun2025efficient}.
With the development of LLMs, recent works \cite{deepseekv4,chu2026kwai} have moved toward hybrid attention that incorporates sparse attention with compressed attention to address the memory-bound nature of decoding.
For example, DeepSeek-V4’s heavily compressed attention \cite{deepseekv4} merges the KV pairs of contiguous tokens within a long span into a single entry, which effectively reduces the KV cache memory footprint.
Our sliding-window compression shares a similar intuition with DeepSeek-V4’s compressed attention: both aim to perform block-level compression to preserve local contextual information while reducing KV cache memory, and DeepSeek-V4 performs KV merging whereas we perform KV eviction.
The key difference lies in the training requirement, in which compressed attention typically requires LLM-coupled pretraining, while our method is training-free and can be applied as a plug-in to a broad range of LLMs.

\section{Preliminary}
\paragraph{Notation.}
We consider an autoregressive Transformer-based LLM \cite{sia2024where,achiam2023gpt} that generates a token sequence denoted by $\mathbf{X}=(x_1,x_2,\dots,x_S)$, where $S$ is the sequence length.
The model consists of $L$ Transformer layers, each with $H$ attention heads and head dimension $d$. 
For the $h$-th attention head, we denote the projected query, key, and value states as $\mathbf{Q}^{l,h},\mathbf{K}^{l,h},\mathbf{V}^{l,h}\in\mathbb{R}^{S\times d}$, respectively.  
In layer $l$ and head $h$, the scaled dot-product attention probabilities are defined as:
\begin{equation}
\mathbf{P}^{l,h}=\text{softmax}(\frac{\mathbf{Q}^{l,h}(\mathbf{K}^{l,h})^\top}{\sqrt{d}} + \mathbf{M}),
\end{equation}
where $\mathbf{P}^{l,h}_{i,j}$ denotes the attention probability from the $i$-th query token to the $j$-th key token, the causal attention mask $\mathbf{M}$ is an upper triangular matrix with nonzero values of $-\infty$.
For simplicity, we use $\mathbf{Q}$ and $\mathbf{K}$ to denote the query and key states of a specific layer $l$ and head $h$.
\paragraph{LLM Inference.}
During autoregressive decoding, the model maintains a key cache $\mathcal{K}$ and a value cache $\mathcal{V}$ to store the key and value states corresponding to prefix tokens.
KV caching avoids recomputing attention over the entire prefix from scratch and improves inference efficiency \cite{li2025a}.

In practice, the decoding efficiency of LLMs is commonly measured by TPOT (Time Per Output Token), which denotes the average time required to generate one output token during decoding and reflects the decoding latency.
Beyond TPOT, throughput is another important metric that measures the number of output tokens generated per unit time, and is commonly used to reflect the concurrency.
In general, more concurrently decoding sequences enable higher throughput.
In long-context scenarios, the inference overhead may be dominated by memory access to the KV cache, leaving the decoding phase memory-bound.
We present the formulas of TPOT and throughput in {Appendix~\ref{appx:tpot}}.

To improve the serving efficiency of LLM inference, frameworks \cite{kwon2023efficient,sglang} such as vLLM have been widely developed, and they aim to maximize memory utilization and sustain efficient decoding under high-concurrency and long-context generation.
The core idea of vLLM is {PagedAttention}, which views the KV cache as a group of fixed-size blocks.
Each block stores a contiguous segment of KV pairs (\eg 256 tokens), and memory allocation and release are managed at the block level during inference.
However, in memory-constrained settings, the available memory can be rapidly consumed as decoding proceeds, which may force the inference framework to reduce the number of running requests and thereby limit overall output throughput (shown in Figure~\ref{fig:F1} (b)).
\paragraph{KV Cache Compression.}
Most compression methods \cite{cai2026rkv,li2024snapkv,yuan-etal-2024-kv,qin2025cake} estimate an importance score $\mathbf{A}_i^{l,h}$ for the $i$-th KV cache entry in attention head $h$ and layer $l$, and then select the most important KV pairs under a cache budget when the cache length reaches a predefined threshold.
A representative score-based compression paradigm can be formulated as:
\begin{equation}
\label{Eq:Topk}
    \mathcal{I}^{l,h} = \textnormal{TopK}(\{\mathbf{A}_i^{l,h}\}_{i=1}^{C_t},\, N)
\end{equation}
\begin{equation}
(\mathcal{K}^{l,h},\mathcal{V}^{l,h})\leftarrow(\mathcal{K}^{l,h},\mathcal{V}^{l,h})[\mathcal{I}^{l,h}],\quad \text{when } C_t \ge C_{\max},
\label{eq:threshold_triggered_compression}
\end{equation}
where $\mathcal{I}^{l,h}$ is the index of selected KV pairs of head $h$ and layer $l$, $N$ is the predefined KV budget, $C_t$ denotes the number of KV pairs currently stored in the cache (\ie the cache length) at decoding step $t$, $C_{\max}$ denotes the predefined threshold.
Once compressed, the cache length is constrained within the interval $[N, C_{\max}]$ as it grows from $N$ back to $C_{\max}$ before the next compression.
Existing methods often set the gap between the budget and the threshold to 128 or 256 to obtain optimal performance \cite{cai2026rkv,devoto2025expectedattention}.
In the following, we omit the notation $l$ and $h$ as all operations performed in the layer and attention head
are the same in our method.
Although the compression paradigm in Equation (\ref{Eq:Topk}) and (\ref{eq:threshold_triggered_compression}) is widely adopted, it suffers from two key limitations:
\begin{itemize}[leftmargin=1.0em]
\item \textbf{Threshold-triggered compression may degrade both reasoning performance and inference efficiency.}
Since the paradigm operates compression over the entire KV cache, it may fully remove all KV pairs within a long contiguous span of context (\eg $x_1 \sim x_{256}$  in Figure \ref{fig:F2}), potentially exacerbating information loss.
Furthermore, when the number of concurrently decoding sequences increases toward the gap between $C_{\max}$ and $N$ (\ie $C_{\max}-N$), compression can be triggered frequently across concurrently decoding sequences, making the overall compression overhead more pronounced and potentially reducing throughput (shown in Figure \ref{fig:F1}).
\item \textbf{Retention granularity is often either isolated KV pairs or rigid fixed-size chunks.} Mainstream score-based methods select isolated KV pairs for retention, which can lead to the loss of semantic information.
Although some methods retain contiguous KV pairs to form chunks, such chunks are often extracted with rigid boundaries (\eg from ${x}_{1}$ to ${x}_{10}$), which may fail to capture flexible-sized cache information distributed at arbitrary positions across the sequence.
\end{itemize}
Given the flaws of existing methods, we argue for using sliding-window bidirectional-attention scores to guide KV-pair retention within the window, together with a retention strategy that incorporates both token-level and chunk-level KV pairs.
We further adapt our method to PagedAttention and develop the {KvLLM} framework with a periodic compression policy, which aims to improve throughput under high concurrency by controlling when compression is triggered.
\section{Method}
\label{sec:method}
\begin{figure}[t]
  \centering
  \begin{minipage}[t]{0.55\textwidth}   
    \centering
    \setlength{\abovecaptionskip}{0pt}
    \setlength{\belowcaptionskip}{-15pt}

    \includegraphics[width=1.1\linewidth, trim=30 20 -30 0]{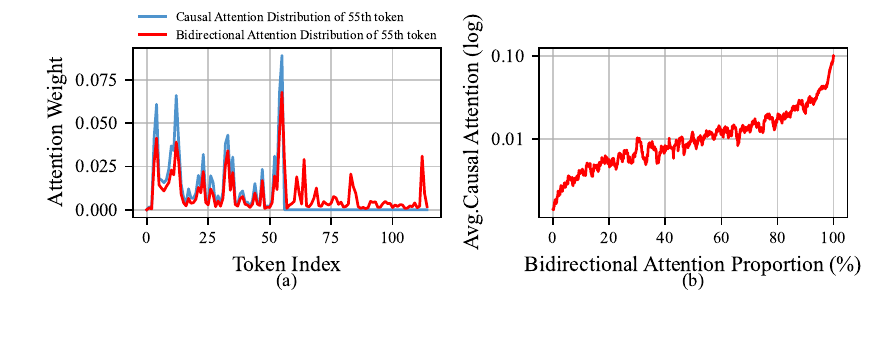}
    \caption{(a) Distribution comparison between causal attention and bidirectional attention of a specific token.
(b) Average causal attention weight versus bidirectional attention percentile.
We first compute the bidirectional attention weights for all token pairs $(x_i, x_j)$ with $j > i$, using $x_i$ as the query and $x_j$ as the key. We then sort these weights in ascending order and group the pairs by percentiles. Finally, for each percentile group, we compute the causal attention weight for the same pairs in the reverse direction ($x_j$ as the query and $x_i$ as the key), average the causal attention weights within the group, and plot the resulting averages against the bidirectional-attention percentiles. 
}
    \label{fig:F3}
  \end{minipage}
  \hfill
  \begin{minipage}[t]{0.40\textwidth}   
    \centering
    \setlength{\abovecaptionskip}{0pt}
    \setlength{\belowcaptionskip}{-15pt}
\includegraphics[width=0.9\linewidth, trim=20 -5 -10 -0]{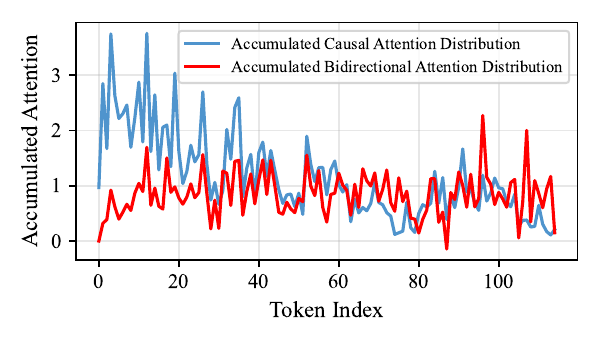}
    \caption{Comparison between accumulated causal attention and bidirectional attention distributions. 
    We compute causal attention and bidirectional attention probability matrices in a given sequence and accumulate the attention received by each token, respectively.}
    \label{fig:F4}
  \end{minipage}
\end{figure}
\subsection{Sliding-Window Bidirectional Attention for KV Importance Estimation}
\label{subsec:method:biwindow}
During decoding, we maintain a sliding window that dynamically covers recently generated tokens, and we apply KV cache compression {only} to KV pairs within the window. 
Once compression is performed, the window slides forward to cover newly generated tokens that are uncompressed, and the same procedure is applied repeatedly. 
Inspired by \cite{cai2026rkv} and \cite{li2024snapkv}, we additionally reserve a buffer at the end of the window to protect the latest generated tokens from being compressed, which helps preserve reasoning quality.

To obtain important {discrete} KV pairs, we first extract the query states of all tokens inside the window and compute {bidirectional attention} to the key states of tokens inside the window except those in the buffer. 
We then accumulate the attention weights received by each KV pair to estimate its importance score. 
Finally, we select the TopK important KV pairs inside the window as retention candidates. 
In summary, the sliding-window bidirectional attention scoring can be formulated as:
\begin{equation}
\label{eq:kara_importance}
\mathbf{A}
=
\operatorname{colsum}(
\operatorname{softmax}(
\frac{\mathbf{Q}_{\mathcal{W}}\mathbf{K}_{\mathcal{W}\setminus\mathcal{U}}^\top}{\sqrt{d}}
)),
\end{equation}
\begin{equation}
\mathcal{I}=\textnormal{TopK}(
\{\mathbf{A}_{j}\}_{j\in \mathcal{W}\setminus\mathcal{U}},
\left\lceil r (|\mathcal{W}|-|\mathcal{U}|)\right\rceil - \alpha).
\label{eq:kara_topk}
\end{equation}
where $\mathcal{W}$ denotes the set of token indices in the current sliding window (\ie $\mathcal{W} = \{1,...,|\mathcal{W}|\}$), and $\mathbf{Q}_{\mathcal{W}}\in\mathbb{R}^{|\mathcal{W}|\times d}$ is the query matrix of the tokens in the sliding window $\mathcal{W}$.
$\mathcal{U}\subset\mathcal{W}$ denotes the buffer at the end of the window (\ie $\mathcal{U} = \{|\mathcal{W}|-|\mathcal{U}|+1,...,|\mathcal{W}|\}$). $\mathcal{W}\setminus\mathcal{U}$ is the compressible region in the window (\ie $\mathcal{W}\setminus\mathcal{U} = \{1,\ldots,|\mathcal{W}|-|\mathcal{U}|\}$). $\mathbf{K}_{\mathcal{W}\setminus\mathcal{U}}\in\mathbb{R}^{|\mathcal{W}\setminus\mathcal{U}|\times d}$ is the key matrix of the tokens in the compressible region $\mathcal{W}\setminus\mathcal{U}$. 
The softmax($\cdot$) is applied over the key dimension and colsum($\cdot$) denotes summing over all queries.
$\mathbf{A}\in\mathbb{R}^{|\mathcal{W}|-|\mathcal{U}|}$ denotes the estimated importance scores of the KV pairs in the compressible region $\mathcal{W}\setminus\mathcal{U}$.
$r\in(0,1]$ is the overall retention ratio of the compressible region, $\alpha$ is the predefined chunk budget controlling the number of additional KV pairs introduced by Token2Chunk, and $\mathcal{I}$ denotes the index set of candidate discrete KV pairs to be retained within the current window.

Since tokens in $\mathcal{U}$ are excluded from compression in the current window, Equation (\ref{eq:kara_importance}) computes bidirectional attention scores using queries from all tokens in $\mathcal{W}$, with scoring restricted to the key states in the compressible region $\mathcal{W}\setminus\mathcal{U}$.
Equation (\ref{eq:kara_topk}) retains candidate KV pairs in the compressible region $\mathcal{W}\setminus\mathcal{U}$.
After compression, the window moves forward by $|\mathcal{W}|-|\mathcal{U}|$ steps once there are $|\mathcal{W}|-|\mathcal{U}|$ newly generated uncompressed tokens beyond the current window.
Implementation details of sliding-window bidirectional attention scoring are presented in Appendix \ref{appx:Implent_SWBA}.
\paragraph{Empirical Analysis.} We conduct simple experiments
to illustrate bidirectional attention in importance estimation. 
Given a sequence, we first compute both the causal attention and bidirectional attention probability matrices, and visualize the cumulative attention received by each token’s key state from all query tokens.
As shown in Figure \ref{fig:F4}, most accumulated causal attention scores concentrate in the prefix region of the window, as earlier positions are accessible to more queries than later ones. 
Then, we visualize the specific token’s bidirectional attention probability distribution over all tokens in the sequence, as well as its causal attention distribution over its preceding tokens. 
As shown in Figure \ref{fig:F3}(a), we observe that the bidirectional attention distribution of the token at the preceding positions largely matches its causal attention distribution.
Finally, we examine the correlation between the bidirectional attention probability from the earlier token $x_i$ to the later token $x_j$ ($j>i$) and the causal attention probability from later token $x_j$ to the earlier token $x_i$. 
In Figure \ref{fig:F3} (b), we observe that when the earlier token $x_i$ assigns a higher bidirectional attention weight to future token $x_j$, token $x_j$ also tends to assign a higher causal attention weight to the earlier token $x_i$.

The above analyses suggest that  bidirectional attention can identify important preceding information and highlight future tokens that assign high attention weight to the preceding tokens. 
These future tokens often contain rich contextual information.
Building on this observation, we use bidirectional attention to identify KV pairs that are both important and informative for the current window (more results are presented in {Appendix \ref{appx:f3}}).
\subsection{Token2Chunk Module}
Building on the discrete KV indices $\mathcal{I}$ obtained from sliding-window bidirectional attention, we introduce a lightweight {Token2Chunk} module to preserve {flexible-sized} contiguous KV pairs at arbitrary positions within the window.
Specifically, Token2Chunk takes as input discrete retained indices $\mathcal{I}$ and the corresponding importance scores $\mathbf{A}\in \mathbb{R}^{|\mathcal{W}|- |\mathcal{U}|}$. 
We first define the chunk budget that controls  the maximum number of additional KV pairs that can be introduced by this module, together with a maximum chunk length. 
We then sort the indices in $\mathcal{I}$ in ascending order and divide the compressible region into chunks by treating every two KV pairs that are consecutive in the sorted index set as the endpoints of a candidate chunk (not adjacent in the original sequence). 
Each candidate chunk indicates a contiguous span, whose interior positions correspond to KV pairs that can be additionally retained as a chunk. 
Next, we filter candidate chunks by the maximum chunk length, and score each remaining chunk using both its length and the importance scores of the two endpoint KV pairs. 
Finally, we preserve all KV pairs inside the selected chunks, and union them with $\mathcal{I}$ to obtain the final retained indices.
For the $k$-th chunk ($1\le k\le|\mathcal{I}|-1 $), Token2Chunk can be formulated as:
\begin{equation}
\label{eq:chunkscore}
\mathbf{R}_k
=
(\mathbf{A}_{\mathcal{I}[k]} + \mathbf{A}_{\mathcal{I}[k+1]}) \times
{(\mathcal{I}[k+1]-\mathcal{I}[k]-1)},
\end{equation}
\begin{equation}
\label{eq:set_topk}
\mathcal{P}
=
\mathrm{TopK}(
\{\, \mathbf{R}_k \,\}_{k = 1}^{|\mathcal{I}|-1},\;
\left\lfloor \frac{\alpha}{\gamma - 2}\right\rfloor), \quad
\text{s.t. }\ \mathcal{I}[k+1]-\mathcal{I}[k]< \gamma 
\end{equation}
\begin{equation}
\label{eq:mergeset}
\hat{\mathcal{I}}
=
\mathcal{I}
\ \cup\
\bigcup_{k\in \mathcal{P}}
\left\{\, j \ \middle|\  \mathcal{I}[k]< j < \mathcal{I}[k+1] \,\right\}.
\end{equation}
where $\mathbf{R}_k$ is the $k$-th chunk's retention score, $\mathcal{I}$ is the discrete retained indices obtained from sliding-window bidirectional attention (sorted in ascending order), and $\mathcal{I}[k]$ denotes the $k$-th smallest element in $\mathcal{I}$. $\mathbf{A}_{\mathcal{I}[k]}$ is the importance score of the KV pair at index $\mathcal{I}[k]$. 
$\gamma$ is the predefined maximum chunk length (including two endpoints) and $\gamma \ge 3$, and $\alpha$ is the predefined chunk budget controlling the number of additional KV pairs introduced by Token2Chunk.  
$\mathcal{P}$ denotes selected chunk indices returned by $\mathrm{TopK}(\cdot)$, and $\hat{\mathcal{I}}$ denotes the final retained KV pair indices after adding all KV pairs inside the selected chunks. 
We then use $\hat{\mathcal{I}}$ to compress the KV cache of the current sliding window:
\begin{equation}
   (\mathcal{K}_{\mathcal{W}\setminus\mathcal{U}},\mathcal{V}_{\mathcal{W}\setminus\mathcal{U}}) \leftarrow (\mathcal{K}_{\mathcal{W}\setminus\mathcal{U}},\mathcal{V}_{\mathcal{W}\setminus\mathcal{U}})[\hat{\mathcal{I}}],
\end{equation}
where $(\mathcal{K}_{\mathcal{W}\setminus\mathcal{U}},\mathcal{V}_{\mathcal{W}\setminus\mathcal{U}})$ denotes the KV cache of the compressible region within the current sliding window.

In Equation (\ref{eq:chunkscore}), the first term measures chunk importance using the importance scores of two consecutive indices in $\mathcal{I}$,
which serve as the start ($\mathcal{I}[{k}]$) and end ($\mathcal{I}[{k+1}]$) positions of the chunk.
The second term captures the chunk length, and a longer chunk preserves more contextual information.
Equation (\ref{eq:set_topk}) selects the chunks with TopK scores under the maximum-length constraint.
Implementation details of Token2Chunk are presented in Appendix \ref{appx:implent_t2k}.
We visualize the retained positions before and after applying Token2Chunk {in Appendix ~\ref{appx:token2chunkres}}.
\subsection{KvLLM: Periodic Compression with PagedAttention}
\label{subsec:method:kvllm}
We adapt Kara to PagedAttention and develop {KvLLM}, an inference framework that supports applying KV cache compression to improve decoding efficiency in memory-constrained environments. 
To mitigate the concurrency-throughput inversion effect observed in Figure \ref{fig:F1}, KvLLM adopts a {periodic compression} policy that is aligned with Kara’s sliding-window design. 
The pseudo-code and implementation details of the periodic policy are shown in {Appendix \ref{appx:pre_code}.}
Generally, KvLLM maintains a global decoding-step counter from the beginning of inference and applies a periodic compression schedule. 
Specifically, we first set the number of decoding steps between two consecutive compression events.
Then, at each scheduled compression, KvLLM selects a subset of running sequences whose trailing PagedAttention blocks are still uncompressed. 
For each selected sequence, KvLLM treats its trailing blocks as a compression window and applies Kara to compress the KV cache within this window.
Finally, KvLLM performs post-processing, including memory compaction and block release. 
By controlling when compression is triggered, periodic compression avoids frequent triggering across concurrent requests.
Under memory-constrained concurrent serving, KvLLM reduces the KV cache footprint and frees memory to decode more sequences.
The overall complexity of Kara is presented in Appendix~\ref{appx:complexity}.

\begin{figure*}[t]
    \centering    
    \setlength{\abovecaptionskip}{0pt} 
    \includegraphics[width=1\linewidth, trim = 63 10 40 0]{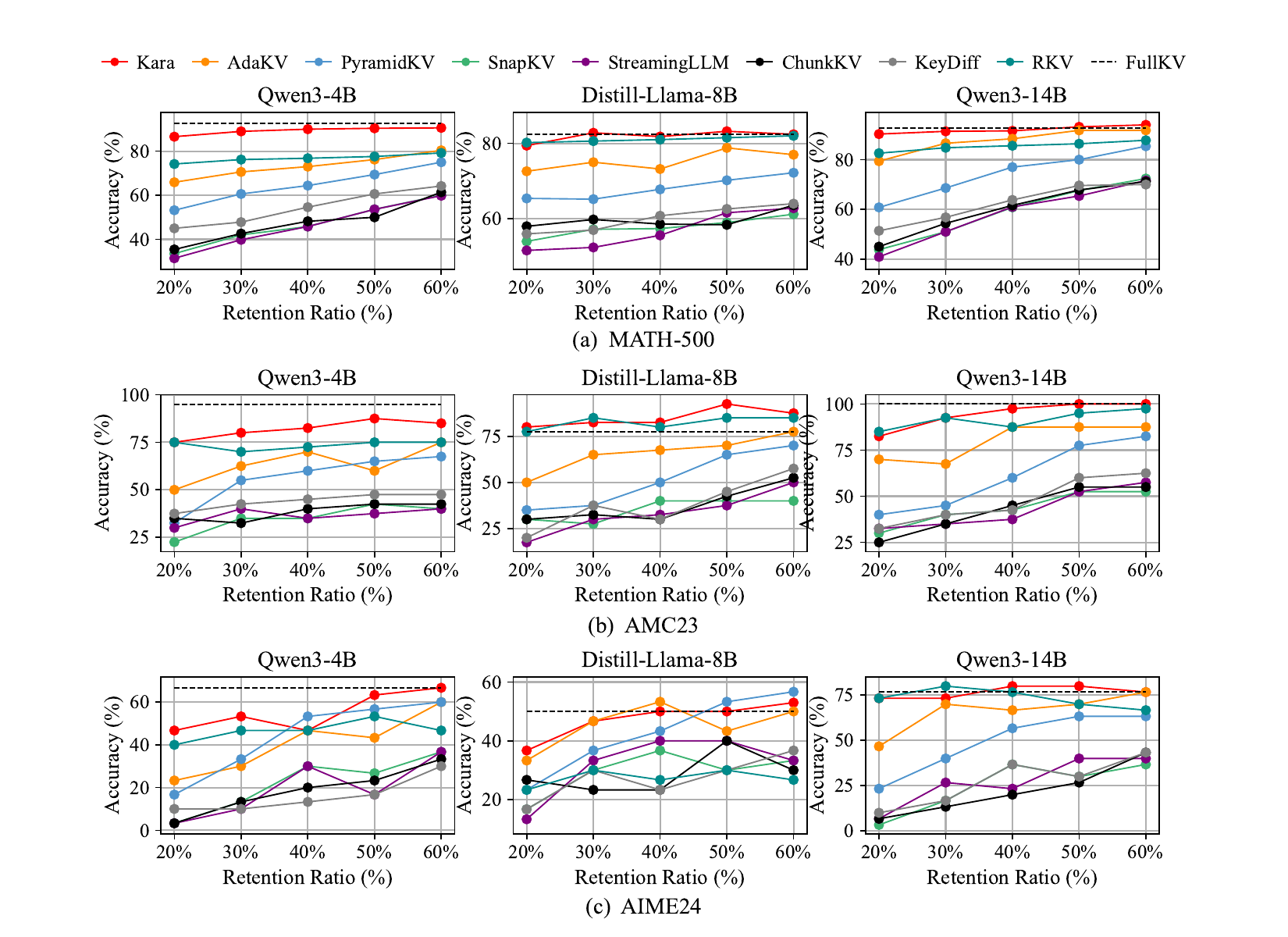}
    \caption{Performance of different KV cache compression methods across varying retention levels. 
    The dash line represents the accuracy of the vanilla LLM model without compression. 
    Note that AdaKV and PyramidKV allocate adaptive memory budgets across layers and heads.}
    \label{fig:sratio}
    \vspace{-10pt}
\end{figure*}
\section{Experiments}
\label{sec:exp}
\paragraph{Benchmarks.}
We evaluate on three mathematical reasoning benchmarks: MATH-500\cite{lin2024criticbench}, AIME 2024\cite{aime24}, and AMC 2023. 
We use greedy decoding and report accuracy of zero-shot pass@1, computed as the fraction of problems answered correctly.
We set the maximum generation length to 16{,}384 tokens for MATH-500 and AMC 2023, and 32{,}768 tokens for AIME 2024. 
We also include a long-context evaluation on the Needle-in-a-Haystack (NIAH) benchmark. 
\paragraph{Baselines.}
We compare Kara against 7 representative decoding-time KV cache compression baselines, including RKV\cite{cai2026rkv}, KeyDiff\cite{park2026keydiff}, SnapKV \cite{li2024snapkv}, StreamingLLM \cite{xiao2024efficient}, PyramidKV \cite{cai2024pyramidkv}, AdaKV \cite{feng2026adakv}, and a chunk-based method ChunkKV \cite{liu2026chunkkv}\footnote{Note that ChunkKV was originally designed for token removal during prefill, and we adapt it to decoding-time compression by treating contiguous KV pairs in the cache as a chunk.}. 
For each baseline, we use its best-performing configuration as recommended by the original paper.
\paragraph{Configurations.}
We conduct experiments on DeepSeek-R1-Distill-Llama-8B (R1-Llama-8B) \cite{guo2025deepseek}, Qwen3-14B, and Qwen3-4B \cite{yang2025qwen3}, using BF16 precision. We evaluate the retention ratio $r\in\{20,30,40,50,60\}\%$, and use the same retention ratio for all layers and attention heads in Kara.
Since Kara does not strictly enforce a fixed compressed cache length, different sequences may result in various cache sizes. 
We ensure a fair comparison by first running Kara to obtain the compressed KV cache length for each sequence and setting the per-sequence KV budget of each baseline to match Kara under the corresponding setting.
Note that we evaluate Kara’s reasoning pass@1 without the periodic compression policy for a fair accuracy comparison, since this policy is primarily designed to improve throughput in practical application, and we only activate it in throughput experiments.
More hyperparameter settings are presented in Appendix \ref{appx:exp_set}.
\subsection{Main Results}
\paragraph{Reasoning Performance.}
Figure~\ref{fig:sratio} reports the reasoning performance of Kara compared with KV cache compression methods under various retention ratios. 
We observe that: 1) Kara preserves accuracy with no or only minor degradation relative to the vanilla LLM without compression. 
For example, on MATH-500 with Qwen3-14B, Kara maintains nearly unchanged accuracy at a 30\% retention ratio. 
2) Kara generally outperforms other compression methods across retention ratios.
3) Kara outperforms AdaKV and PyramidKV in most cases, which allocate different memory budgets across layers and heads.
In contrast, we use a unified configuration and retention ratio for all layers and heads, which is architecturally more compatible with tensor parallelism and PagedAttention in inference frameworks.
We further discuss these results in Appendix \ref{appx:dis_mainres}.
\paragraph{NIAH Performance.} To evaluate Kara’s ability to preserve context information under KV cache compression, we conduct a Needle-in-a-Haystack (NIAH) experiment. 
The results are presented in Appendix \ref{appx:exp_niah}.
We observe that Kara achieves higher retrieval accuracy than the baselines across most depths and context lengths. 
\begin{table}  
\centering
\small
\setlength{\abovecaptionskip}{0.1cm}
\caption{Ablation study of Kara with Qwen3-4B at a 40\% KV retention ratio.}
\setlength{\tabcolsep}{0.008\textwidth}
\begin{tabular}{c|cc}
\toprule
{Datasets} & MATH-500 & AMC23\\
\midrule
w/o Bi-Attention & 85.20 & 72.50\\
w/o Token2Chunk & 88.20 & 77.50\\
Kara & \textbf{90.00} & \textbf{82.50} \\
\bottomrule
\end{tabular}
\label{tab:ab}
\end{table}
\subsection{Ablation Studies}
\paragraph{Effects of Bidirectional Attention Scoring.}
To evaluate the effectiveness of the sliding-window bidirectional attention scoring in Equation (\ref{eq:kara_importance}), we remove bidirectional attention and score KV pairs in the compressible region using only the query states from the protected buffer $\mathcal{U}$ (denoted as w/o Bi-Attention). 
As shown in Table~\ref{tab:ab}, removing bidirectional attention consistently reduces reasoning accuracy. 
These results indicate that utilizing all query states in the window with bidirectional attention provides a stronger importance signal than buffer-only causal attention scoring, leading to better preservation of important and informative context.
\paragraph{Effects of Token2Chunk.}
To assess the effectiveness of the Token2Chunk module in Equation (\ref{eq:chunkscore}), we remove Token2Chunk and retain KV pairs using only bidirectional attention scoring, selecting a target number of discrete KV pairs under the same retention ratio (w/o Token2Chunk). As shown in Table~\ref{tab:ab}, removing Token2Chunk leads to lower performance under compression. 
These results show that preserving flexible-sized chunks at arbitrary positions helps preserve more semantic context.
\begin{table}[t]
\vspace{-0pt}
\centering
\caption{Overall inference latency and average throughput under the predefined maximum running sequences.}
\label{tab:batch_results_128_256}
\renewcommand{\arraystretch}{1.1}
\setlength{\tabcolsep}{4pt}
\small
\begin{tabular}{l|cc|cc}
\toprule
\textbf{}
  & \multicolumn{2}{c|}{128}
  & \multicolumn{2}{c}{256} \\
\midrule
\textbf{Metric}
  & Latency (s) & Throughput (tokens/s)
  & Latency (s) & Throughput (tokens/s) \\
\midrule
vLLM
  & 884  & 2,633
  & 678  & 2,943 \\
vLLM + SnapKV
  & 920  & 2,341
  & 738  & 2,556 \\
KvLLM
  & \textbf{864}  & \textbf{2,912}
  & \textbf{640}  & \textbf{3,392} \\
\bottomrule
\end{tabular}
\vspace{-10pt}
\end{table}
\subsection{KvLLM Throughput and Latency Evaluation.}
To evaluate the efficiency benefits of Kara in practical serving, we adapt it to PagedAttention and develop the KvLLM inference framework. 
We report both average throughput and overall inference latency for KvLLM, vanilla vLLM, and vLLM equipped with SnapKV under the predefined maximum number of concurrently decoding sequences 128 and 256.
For a fair comparison, we simulate a memory-constrained environment for all frameworks, and the detailed settings are presented in Appendix \ref{appx:vllm_set}.
The results are shown in Table \ref{tab:batch_results_128_256}. 
We observe that KvLLM with periodic compression improves output throughput and reduces decoding latency across these settings. 
The improvement can be attributed to the PagedAttention blocks freed by each compression. 
These freed blocks can then be used to decode other waiting sequences, thereby increasing throughput and reducing latency.
In contrast, the threshold-triggered policy of vLLM+SnapKV can trigger compression frequently. In addition, the accumulated overhead of query buffer maintenance and global memory compaction outweighs the throughput gain from compression.
Instead, KvLLM controls when compression is triggered and recomputes query states within the window.
It compacts cache memory only in the window region.
Overall, these results show that KvLLM, which combines periodic compression with sliding-window compression, effectively improves throughput and reduces latency under memory constraints.
Additional discussion and results on more batch sizes, and the actual concurrency across decoding steps are provided in Appendix \ref{appx:kvllm}.
\section{Conclusion}
We propose Kara, a decoding-time sliding-window KV cache compression method that scores and retains KV pairs in the recent context using bidirectional attention and augments discrete retention with a Token2Chunk module for flexible-sized  chunks.
We further integrate it into  PagedAttention and develop the KvLLM framework with periodic trailing-block compression to reduce KV-cache memory usage and improve throughput.
In the future, we plan to combine KV offloading and retrieval to enable lossless KV cache compression.
\section{Acknowledgement}
We thank Yang Zhou and Beidi Chen group for their helpful discussions, valuable feedback, and support throughout this project.  

\bibliographystyle{unsrt}
\bibliography{sample-base}

@article{wei2022chain,
  title={Chain-of-thought prompting elicits reasoning in large language models},
  author={Wei, Jason and Wang, Xuezhi and Schuurmans, Dale and Bosma, Maarten and Xia, Fei and Chi, Ed and Le, Quoc V and Zhou, Denny and others},
  journal={Advances in neural information processing systems},
  volume={35},
  pages={24824--24837},
  year={2022}
}

@article{yang2025qwen3,
  title={Qwen3 technical report},
  author={Yang, An and Li, Anfeng and Yang, Baosong and Zhang, Beichen and Hui, Binyuan and Zheng, Bo and Yu, Bowen and Gao, Chang and Huang, Chengen and Lv, Chenxu and others},
  journal={arXiv preprint arXiv:2505.09388},
  year={2025}
}

@article{li2024survey,
  title={A survey on large language model acceleration based on kv cache management},
  author={Li, Haoyang and Li, Yiming and Tian, Anxin and Tang, Tianhao and Xu, Zhanchao and Chen, Xuejia and Hu, Nicole and Dong, Wei and Li, Qing and Chen, Lei},
  journal={arXiv preprint arXiv:2412.19442},
  year={2024}
}

@article{yuan2024llm,
  title={Llm inference unveiled: Survey and roofline model insights},
  author={Yuan, Zhihang and Shang, Yuzhang and Zhou, Yang and Dong, Zhen and Zhou, Zhe and Xue, Chenhao and Wu, Bingzhe and Li, Zhikai and Gu, Qingyi and Lee, Yong Jae and others},
  journal={arXiv preprint arXiv:2402.16363},
  year={2024}
}

@inproceedings{
cai2026rkv,
title={R-{KV}: Redundancy-aware {KV} Cache Compression for Reasoning Models},
author={Zefan Cai and Wen Xiao and Hanshi Sun and Cheng Luo and Yikai Zhang and Ke Wan and Yucheng Li and Yeyang Zhou and Li-Wen Chang and Jiuxiang Gu and Zhen Dong and Anima Anandkumar and Abedelkadir Asi and Junjie Hu},
booktitle={The Thirty-ninth Annual Conference on Neural Information Processing Systems},
year={2026},
url={https://openreview.net/forum?id=2jwAjomEDB}
}

@article{guo2025deepseek,
  title={Deepseek-r1: Incentivizing reasoning capability in llms via reinforcement learning},
  author={Guo, Daya and Yang, Dejian and Zhang, Haowei and Song, Junxiao and Wang, Peiyi and Zhu, Qihao and Xu, Runxin and Zhang, Ruoyu and Ma, Shirong and Bi, Xiao and others},
  journal={arXiv preprint arXiv:2501.12948},
  year={2025}
}

@inproceedings{
li2024snapkv,
title={Snap{KV}: {LLM} Knows What You are Looking for Before Generation},
author={Yuhong Li and Yingbing Huang and Bowen Yang and Bharat Venkitesh and Acyr Locatelli and Hanchen Ye and Tianle Cai and Patrick Lewis and Deming Chen},
booktitle={The Thirty-eighth Annual Conference on Neural Information Processing Systems},
year={2024},
url={https://openreview.net/forum?id=poE54GOq2l}
}

@inproceedings{
feng2026adakv,
title={Ada-{KV}: Optimizing {KV} Cache Eviction by Adaptive Budget Allocation for Efficient {LLM} Inference},
author={Yuan Feng and Junlin Lv and Yukun Cao and Xike Xie and S Kevin Zhou},
booktitle={The Thirty-ninth Annual Conference on Neural Information Processing Systems},
year={2026},
url={https://openreview.net/forum?id=tcisuhGsQZ}
}

@inproceedings{
xiao2024efficient,
title={Efficient Streaming Language Models with Attention Sinks},
author={Guangxuan Xiao and Yuandong Tian and Beidi Chen and Song Han and Mike Lewis},
booktitle={The Twelfth International Conference on Learning Representations},
year={2024},
url={https://openreview.net/forum?id=NG7sS51zVF}
}

@inproceedings{zhang2025beyond,
  title={Beyond text-visual attention: Exploiting visual cues for effective token pruning in vlms},
  author={Zhang, Qizhe and Cheng, Aosong and Lu, Ming and Zhang, Renrui and Zhuo, Zhiyong and Cao, Jiajun and Guo, Shaobo and She, Qi and Zhang, Shanghang},
  booktitle={Proceedings of the IEEE/CVF International Conference on Computer Vision},
  pages={20857--20867},
  year={2025}
}

@inproceedings{
dms,
title={Inference-Time Hyper-Scaling with {KV} Cache Compression},
author={Adrian {\L}a{\'n}cucki and Konrad Staniszewski and Piotr Nawrot and Edoardo Ponti},
booktitle={The Thirty-ninth Annual Conference on Neural Information Processing Systems},
year={2026},
url={https://openreview.net/forum?id=8ZiElzQxf1}
}

@inproceedings{
kim2026kvzip,
title={{KV}zip: Query-Agnostic {KV} Cache Compression with Context Reconstruction},
author={Jang-Hyun Kim and Jinuk Kim and Sangwoo Kwon and Jae W. Lee and Sangdoo Yun and Hyun Oh Song},
booktitle={The Thirty-ninth Annual Conference on Neural Information Processing Systems},
year={2026},
url={https://openreview.net/forum?id=JFygzwx8SJ}
}

@article{devoto2025expectedattention,
  title={Expected Attention: KV Cache Compression by Estimating Attention from Future Queries Distribution},
  author={Devoto, Alessio and Jeblick, Maximilian and J{\'e}gou, Simon},
  journal={arXiv preprint arXiv:2510.00636},
  year={2025},
  url={https://arxiv.org/abs/2510.00636}
}

@inproceedings{
liu2026chunkkv,
title={Chunk{KV}: Semantic-Preserving {KV} Cache Compression for Efficient Long-Context {LLM} Inference},
author={Xiang Liu and Zhenheng Tang and Peijie Dong and Zeyu Li and Liuyue and Bo Li and Xuming Hu and Xiaowen Chu},
booktitle={The Thirty-ninth Annual Conference on Neural Information Processing Systems},
year={2026},
url={https://openreview.net/forum?id=20JDhbJqn3}
}

@article{achiam2023gpt,
  title={Gpt-4 technical report},
  author={Achiam, Josh and Adler, Steven and Agarwal, Sandhini and Ahmad, Lama and Akkaya, Ilge and Aleman, Florencia Leoni and Almeida, Diogo and Altenschmidt, Janko and Altman, Sam and Anadkat, Shyamal and others},
  journal={arXiv preprint arXiv:2303.08774},
  year={2023}
}

@inproceedings{
sia2024where,
title={Where does In-context  Learning {\textbackslash}{\textbackslash} Happen in Large Language Models?},
author={Suzanna Sia and David Mueller and Kevin Duh},
booktitle={The Thirty-eighth Annual Conference on Neural Information Processing Systems},
year={2024},
url={https://openreview.net/forum?id=LLuSjg59an}
}

@inproceedings{kwon2023efficient,
  title={Efficient Memory Management for Large Language Model Serving with PagedAttention},
  author={Woosuk Kwon and Zhuohan Li and Siyuan Zhuang and Ying Sheng and Lianmin Zheng and Cody Hao Yu and Joseph E. Gonzalez and Hao Zhang and Ion Stoica},
  booktitle={Proceedings of the ACM SIGOPS 29th Symposium on Operating Systems Principles},
  year={2023}
}

@inproceedings{lin2024criticbench,
  title={Criticbench: Benchmarking llms for critique-correct reasoning},
  author={Lin, Zicheng and Gou, Zhibin and Liang, Tian and Luo, Ruilin and Liu, Haowei and Yang, Yujiu},
  booktitle={Findings of the Association for Computational Linguistics: ACL 2024},
  pages={1552--1587},
  year={2024}
}

@article{grattafiori2024llama,
  title={The llama 3 herd of models},
  author={Grattafiori, Aaron and Dubey, Abhimanyu and Jauhri, Abhinav and Pandey, Abhinav and Kadian, Abhishek and Al-Dahle, Ahmad and Letman, Aiesha and Mathur, Akhil and Schelten, Alan and Vaughan, Alex and others},
  journal={arXiv preprint arXiv:2407.21783},
  year={2024}
}

@article{
li2025a,
title={A Survey on Large Language Model Acceleration based on {KV} Cache Management},
author={Haoyang LI and Yiming Li and Anxin Tian and Tianhao Tang and Zhanchao Xu and Xuejia Chen and Nicole HU and Wei Dong and Li Qing and Lei Chen},
journal={Transactions on Machine Learning Research},
issn={2835-8856},
year={2025},
url={https://openreview.net/forum?id=z3JZzu9EA3},
note={}
}

@inproceedings{sglang,
author = {Zheng, Lianmin and Yin, Liangsheng and Xie, Zhiqiang and Sun, Chuyue and Huang, Jeff and Yu, Cody Hao and Cao, Shiyi and Kozyrakis, Christos and Stoica, Ion and Gonzalez, Joseph E. and Barrett, Clark and Sheng, Ying},
title = {SGLang: efficient execution of structured language model programs},
year = {2024},
isbn = {9798331314385},
publisher = {Curran Associates Inc.},
address = {Red Hook, NY, USA},
articleno = {2000},
numpages = {27},
location = {Vancouver, BC, Canada},
series = {NIPS '24}
}

@inproceedings{
ramachandran2026thinkv,
title={Thin{KV}: Thought-Adaptive {KV} Cache Compression for Efficient Reasoning Models},
author={Akshat Ramachandran and Marina Neseem and Charbel Sakr and Rangharajan Venkatesan and Brucek Khailany and Tushar Krishna},
booktitle={The Fourteenth International Conference on Learning Representations},
year={2026},
url={https://openreview.net/forum?id=M3CeHnZKNC}
}

@inproceedings{
park2026keydiff,
title={KeyDiff: Key Similarity-Based {KV} Cache Eviction for Long-Context {LLM} Inference in Resource-Constrained Environments},
author={Junyoung Park and Dalton Jones and Matthew J Morse and Raghavv Goel and Mingu Lee and Christopher Lott},
booktitle={The Thirty-ninth Annual Conference on Neural Information Processing Systems},
year={2026},
url={https://openreview.net/forum?id=uBaFH7aQnC}
}

@misc{aime24,
      title={American Invitational Mathematics Examination (AIME) 2024}, 
      author={Zhang, Yifan and Math-AI, Team},
      year={2024},
}

@inproceedings{
jones2026quoka,
title={Quo{KA}: Query-Oriented {KV} Selection for Efficient {LLM} Prefill},
author={Dalton Jones and Junyoung Park and Matthew J Morse and Mingu Lee and Matthew Harper Langston and Christopher Lott},
booktitle={The Fourteenth International Conference on Learning Representations},
year={2026},
url={https://openreview.net/forum?id=YS4N1YxXSM}
}

@inproceedings{
mao2026icecache,
title={IceCache: Memory-Efficient {KV}-cache Management for Long-Sequence {LLM}s},
author={Yuzhen Mao and Qitong Wang and Martin Ester and Ke Li},
booktitle={The Fourteenth International Conference on Learning Representations},
year={2026},
url={https://openreview.net/forum?id=yHxSKM9kdr}
}

@article{cai2024pyramidkv,
  title={Pyramidkv: Dynamic kv cache compression based on pyramidal information funneling},
  author={Cai, Zefan and Zhang, Yichi and Gao, Bofei and Liu, Yuliang and Li, Yucheng and Liu, Tianyu and Lu, Keming and Xiong, Wayne and Dong, Yue and Hu, Junjie and others},
  journal={arXiv preprint arXiv:2406.02069},
  year={2024}
}

@inproceedings{yuan-etal-2024-kv,
    title = "{KV} Cache Compression, But What Must We Give in Return? A Comprehensive Benchmark of Long Context Capable Approaches",
    author = "Yuan, Jiayi  and
      Liu, Hongyi  and
      Zhong, Shaochen  and
      Chuang, Yu-Neng  and
      Li, Songchen  and
      Wang, Guanchu  and
      Le, Duy  and
      Jin, Hongye  and
      Chaudhary, Vipin  and
      Xu, Zhaozhuo  and
      Liu, Zirui  and
      Hu, Xia",
    editor = "Al-Onaizan, Yaser  and
      Bansal, Mohit  and
      Chen, Yun-Nung",
    booktitle = "Findings of the Association for Computational Linguistics: EMNLP 2024",
    month = nov,
    year = "2024",
    address = "Miami, Florida, USA",
    publisher = "Association for Computational Linguistics",
    url = "https://aclanthology.org/2024.findings-emnlp.266/",
    doi = "10.18653/v1/2024.findings-emnlp.266",
    pages = "4623--4648",
}

@inproceedings{
qin2025cake,
title={{CAKE}: Cascading and Adaptive {KV} Cache Eviction with Layer Preferences},
author={Ziran Qin and Yuchen Cao and Mingbao Lin and Wen Hu and Shixuan Fan and Ke Cheng and Weiyao Lin and Jianguo Li},
booktitle={The Thirteenth International Conference on Learning Representations},
year={2025},
url={https://openreview.net/forum?id=EQgEMAD4kv}
}

@inproceedings{
trimkv,
title={Cache What Lasts: Token Retention for Memory-Bounded {KV} Cache in {LLM}s},
author={Ngoc Bui and Shubham Sharma and Simran Lamba and Saumitra Mishra and Rex Ying},
booktitle={The Fourteenth International Conference on Learning Representations},
year={2026},
url={https://openreview.net/forum?id=qCaq3jGb0S}
}

@misc{deepseekv4,
      title={DeepSeek-V4: Towards Highly Efficient Million-Token Context Intelligence},
      author={DeepSeek-AI},
      year={2026},
}

@inproceedings{
zhang2025spargeattention,
title={SpargeAttention: Accurate and Training-free Sparse Attention Accelerating Any Model Inference},
author={Jintao Zhang and Chendong Xiang and Haofeng Huang and Jia wei and Haocheng Xi and Jun Zhu and Jianfei Chen},
booktitle={Forty-second International Conference on Machine Learning},
year={2025},
url={https://openreview.net/forum?id=74c3Wwk8Tc}
}

@inproceedings{
lu2026moba,
title={Mo{BA}: Mixture of Block Attention for Long-Context {LLM}s},
author={Enzhe Lu and Zhejun Jiang and Jingyuan Liu and Yulun Du and Tao Jiang and Chao Hong and Shaowei Liu and Weiran He and Enming Yuan and Yuzhi Wang and Zhiqi Huang and Huan Yuan and Suting Xu and Xinran Xu and Guokun Lai and Yanru Chen and Huabin Zheng and Junjie Yan and Jianlin Su and Yuxin Wu and Yutao Zhang and Zhilin Yang and Xinyu Zhou and Mingxing Zhang and Jiezhong Qiu},
booktitle={The Thirty-ninth Annual Conference on Neural Information Processing Systems},
year={2026},
url={https://openreview.net/forum?id=RlqYCpTu1P}
}

@inproceedings{
lin2026twilight,
title={Twilight: Adaptive Attention Sparsity with Hierarchical Top-\$p\$  Pruning},
author={Chaofan Lin and Jiaming Tang and Shuo Yang and Hanshuo Wang and Tian Tang and Boyu Tian and Ion Stoica and Song Han and Mingyu Gao},
booktitle={The Thirty-ninth Annual Conference on Neural Information Processing Systems},
year={2026},
url={https://openreview.net/forum?id=Ve693NkzcU}
}

@inproceedings{xattn,
  author       = {Ruyi Xu and
                  Guangxuan Xiao and
                  Haofeng Huang and
                  Junxian Guo and
                  Song Han},
  editor       = {Aarti Singh and
                  Maryam Fazel and
                  Daniel Hsu and
                  Simon Lacoste{-}Julien and
                  Felix Berkenkamp and
                  Tegan Maharaj and
                  Kiri Wagstaff and
                  Jerry Zhu},
  title        = {XAttention: Block Sparse Attention with Antidiagonal Scoring},
  booktitle    = {Forty-second International Conference on Machine Learning, {ICML}
                  2025, Vancouver, BC, Canada, July 13-19, 2025},
  series       = {Proceedings of Machine Learning Research},
  publisher    = {{PMLR} / OpenReview.net},
  year         = {2025},
  url          = {https://proceedings.mlr.press/v267/xu25ag.html},
  bibsource    = {dblp computer science bibliography, https://dblp.org}
}

@article{chu2026kwai,
  title={Kwai Summary Attention Technical Report},
  author={Chu, Chenglong and Zhou, Guorui and Zhang, Guowang and Li, Han and Peng, Hao and Cheng, Hongtao and Liang, Jian and Cao, Jiangxia and Gai, Kun and Zhou, Lingzhi and others},
  journal={arXiv preprint arXiv:2604.24432},
  year={2026}
}

@article{sun2025efficient,
  title={Efficient attention mechanisms for large language models: A survey},
  author={Sun, Yutao and Li, Zhenyu and Zhang, Yike and Pan, Tengyu and Dong, Bowen and Guo, Yuyi and Wang, Jianyong},
  journal={arXiv preprint arXiv:2507.19595},
  year={2025}
}

@inproceedings{blocksparse,
  author       = {Emily Xiao and
                  Chin{-}Jou Li and
                  Yilin Zhang and
                  Graham Neubig and
                  Amanda Bertsch},
  editor       = {Wanxiang Che and
                  Joyce Nabende and
                  Ekaterina Shutova and
                  Mohammad Taher Pilehvar},
  title        = {Efficient Many-Shot In-Context Learning with Dynamic Block-Sparse
                  Attention},
  booktitle    = {Proceedings of the 63rd Annual Meeting of the Association for Computational
                  Linguistics (Volume 1: Long Papers), {ACL} 2025, Vienna, Austria,
                  July 27 - August 1, 2025},
  pages        = {31946--31958},
  publisher    = {Association for Computational Linguistics},
  year         = {2025},
  url          = {https://aclanthology.org/2025.acl-long.1542/},
  bibsource    = {dblp computer science bibliography, https://dblp.org}
}

@inproceedings{
banerjee2025crane,
title={{CRANE}: Reasoning with constrained {LLM} generation},
author={Debangshu Banerjee and Tarun Suresh and Shubham Ugare and Sasa Misailovic and Gagandeep Singh},
booktitle={Forty-second International Conference on Machine Learning},
year={2025},
url={https://openreview.net/forum?id=wKs9fHYxCV}
}

@inproceedings{niah,
author = {Fu, Yao and Panda, Rameswar and Niu, Xinyao and Yue, Xiang and Hajishirzi, Hannaneh and Kim, Yoon and Peng, Hao},
title = {Data engineering for scaling language models to 128K context},
year = {2024},
publisher = {JMLR.org},
articleno = {564},
numpages = {10},
location = {Vienna, Austria},
series = {ICML'24}
}


\appendix
\section{Limitations, Future Work and Impact}
\paragraph{Limitations.}
We acknowledge several limitations of Kara and KvLLM.
First, permanently evicting KV pairs inevitably causes information loss and may degrade reasoning performance in broader application settings.
Second, obtaining query states via recomputation is straightforward to implement in practice but introduces non-trivial overhead. Although the periodic compression policy can reduce the frequency of recomputation and improve the overall trade-off, the overhead remains a key challenge.
Third, compared with threshold-triggered policies, Kara and KvLLM do not precisely control the memory footprint, which may be less suitable for settings with strict memory requirements.
Finally, our compression is not global, which may leave redundancy in the KV cache that cannot be fully eliminated.
\paragraph{Future Work.}
In future work, we plan to incorporate KV cache offloading and retrieval into KvLLM. 
Concretely, we view a sequence’s KV cache as two parts: KV pairs that remain persistently important over a long horizon, and KV pairs that are only important in the short term. For each sequence, we will additionally maintain a dedicated block to store short-term KV pairs, while the remaining blocks store long-term KV pairs together with the most recently generated KV pairs.
For compression, we will continue to use periodic compression with two schedules: a long period for updating the long-term blocks and a short period for updating the short-term block. At each compression event, evicted KV pairs will be moved to CPU memory or SSD, and we will use the latest tokens’ query states to retrieve the TopK KV pairs back into HBM for decoding. This offloading-and-retrieval design can mitigate information loss caused by permanent eviction and improve throughput while reducing GPU-memory usage.

To reduce compression overhead, we will also consider training-based approaches to predict importance scores. To make learned scoring effective across broader settings, such training may need to be integrated with sparse-attention modeling, where importance signals can be learned jointly during pretraining. This direction may enable low-overhead KV cache compression that generalizes to more diverse scenarios.

\paragraph{Broader Impacts.}
Kara’s sliding-window compression mechanism and KvLLM’s periodic compression policy provide a practical path for deploying and integrating KV cache compression into inference frameworks.

\section{Detailed Experimental Settings}
\subsection{Detailed Settings for vLLM and KvLLM}
\label{appx:vllm_set}
For all frameworks, to simulate a memory-constrained environment, we use DeepSeek-R1-Distill-LLaMA-8B and the MATH-500 dataset (average generation length is approximately 3,000 tokens). 
We set the tensor-parallel size to 4 and the GPU memory utilization to 0.5 (\ie 16GB). 
The maximum sequence generation length is set to 8,192 tokens. 
We set the block size to 256.

For vLLM+SnapKV, we set the compression threshold to 1280 KV pairs (5 blocks) and set the compressed cache length to 1,025 KV pairs (also 5 blocks). 
The SnapKV query cache length is 32.

For Kara in KvLLM, we set the window length to 3 PagedAttention blocks (768 KV pairs) and the total retained length per window to 257 KV pairs. 
The extra KV pair is reserved to avoid frequent block allocation.
The compression period is set to 128 decoding steps. At each compression, KvLLM selects 30 sequences whose trailing context contains at least 768 uncompressed tokens and applies Kara within the trailing window.
The buffer length is also 32.

All experiments are conducted on 8 RTX 5090 GPUs.
\subsection{Detailed Experimental Settings for Figure \ref{fig:F3} and \ref{fig:F4}}
\label{app:fig34_settings}
This section provides detailed settings for the simple experiment shown in Figure \ref{fig:F3} and \ref{fig:F4}. 
We conduct this analysis using DeepSeek-R1-Distill-Llama-8B. 
We sample multiple prompts from the MATH-500 dataset and extract the Q/K/V states of the first 120 tokens.
We report a representative example. When analyzing attention, we visualize the attention probability distribution averaged over all attention heads for each token.
 

\subsection{Detailed Experimental Settings in Section \ref{sec:exp}}
\label{appx:exp_set}
Experiments for Kara are conducted on a cluster with 8 H200 GPUs and 8 H100 GPUs.
For Kara, we sweep the window length $|\mathcal{W}|\in\{256,384,512\}$ and set the buffer length $|\mathcal{U}|\in\{32,64\}$. 
For Token2Chunk, we use a fixed maximum chunk size $\gamma$ of $8$ and set the additional chunk budget $\alpha$ to $16$ or $32$. 
\section{Details}
\subsection{Formula of TPOT and Throughput}
\label{appx:tpot}
In a simplified decoding setting, where the number of concurrently decoding sequences is denoted by $B$\footnote{$B$ varies dynamically during decoding}, the KV cache uses BF16 precision, and the FFN intermediate size is four times the hidden size, the TPOT can be approximated as follows:
\begin{equation}
\label{Eq:TPOT}
\text{TPOT} \approx \max \left(
\frac{2L(12BD^2 + 2BSD)}{\rho},
\frac{4LBSD}{\beta}
\right),
\end{equation}
where $\rho$ denotes the effective GPU compute throughput, $D = Hd$ is the hidden dimension and $\beta$ denotes the effective HBM bandwidth, and modern GPUs generally have a high compute-to-bandwidth ratio.
The first term in Equation (\ref{Eq:TPOT}) corresponds to the computation cost of one decoding forward step, while the second term captures the memory access overhead of KV cache.
We omit the memory footprint of model weights since we focus on the KV cache.
The output throughput at each decoding step can be formulated as:
\begin{equation}
\label{Eq:throughput}
\text{Throughput} \approx \frac{B}{\text{TPOT}},
\end{equation}
From Equation (\ref{Eq:TPOT}) and \ref{Eq:throughput}, we can observe that in long-context reasoning scenarios, the inference overhead may be dominated by memory access to the KV cache, leaving the inference memory-bound, and higher $B$ generally enables higher throughput.
\subsection{Implementation Details of Sliding-window Bidirectional Attention Scoring}
\label{appx:Implent_SWBA}
In this section, we present the implementation details of sliding-window bidirectional attention.
We can obtain the query states required for KV scoring by recomputing the current sliding window while reusing the existing KV cache.
Specifically, at each compression, in addition to Kara’s computation, we run an additional forward pass on the tokens in the current window. 
In this pass, we compute only the query projections, the causal attention against the cached prefix KV states, and the FFN (feed-forward network), while reusing the existing key and value caches. This recomputation avoids maintaining an explicit query cache during decoding. 
The sliding-window mechanism can also be applied to compress the prompt KV cache {before} decoding. 
Specifically, during the prefill stage, we can partition the prompt into multiple windows of length $|\mathcal{W}|$ and perform window-wise compression on the KV pairs corresponding to positions inside each window in parallel. 
After compression, the window moves forward by $|\mathcal{W}|-|\mathcal{U}|$ steps once there are $|\mathcal{W}|-|\mathcal{U}|$ uncompressed tokens beyond the current window during decoding.
\subsection{Implementation Details of Token2Chunk}
\label{appx:implent_t2k}
During decoding, Token2Chunk is applied to the compressible region immediately after the bidirectional-attention scoring stage.
Token2Chunk is a lightweight module, and 
its additional complexity is $O(n)$, where $n = |\mathcal{I}|$.
Equation (\ref{eq:chunkscore}) and Equation (\ref{eq:set_topk}) ensure that the number of additional KV pairs is bounded by the chunk budget $\alpha$.
If the selected chunks introduce fewer than $\alpha$ additional KV pairs, we fill the chunk budget by selecting discrete KV pairs from the unselected positions in $\mathcal{W}\setminus\mathcal{U}$ according to score $\mathbf{A}_i$.
\subsection{Pseudo Code and Implementation Details of the Periodic Compression Policy}
\label{appx:pre_code}
The pseudocode of the periodic compression policy is presented in Algorithm \ref{alg:kvllm_periodic}.
In our implementation with PagedAttention, the last block of a sequence is not always full. 
We therefore select sequences whose tail contains full and uncompressed blocks corresponding to $\mathcal{W}$ tokens to form the compression window, and we exclude the last (possibly partial) block from compression. 
Since all selected sequences are compressed with the same window size, the blocks processed per sequence have the same shape. 
This allows us to process the selected sequences’ windows in parallel during a compression event, including KV importance estimation, window-level memory compaction.
To guarantee that we can always select sequences with $|\mathcal{W}|$ uncompressed tokens at the tail,
we maintain a variable for each sequence that tracks the current number of uncompressed tokens at its end;
this value is reset to 0 after each compression step.
\begin{algorithm}[t]
\caption{Periodic compression policy of KvLLM}
\label{alg:kvllm_periodic}
\begin{algorithmic}[1]
\Require Running queue $\mathcal{Q}$, empty key/value caches $\mathcal{K}, \mathcal{V}$, global step counter $g \gets 0$, compression period $\delta$, compression window length $|\mathcal{W}|$, buffer length $|\mathcal{U}|$, retention ratio $r$, $\mathrm{Kara}(\cdot)$ (returns retained indices for the compressible region in a trailing window)
\While{$\mathcal{Q}\neq\emptyset$}
    \State Run one decoding step for sequences in $\mathcal{Q}$; update $\mathcal{K},\mathcal{V}$
    \State $g \gets g + 1$
    \If{$g>|\mathcal{W}|$ \textbf{and} $g \bmod \delta = 0$}
        \State $\mathcal{Q}' \gets \textsc{Select}(\mathcal{Q})$ \Comment{select sequences whose trailing $|\mathcal{W}|$ tokens are uncompressed}
 \ForAll{$e \in \mathcal{Q}'$}
    \State $\mathcal{K}_{\mathcal{W}} \gets \mathcal{K}[e,\,-|\mathcal{W}|:]$
    \State $\mathcal{V}_{\mathcal{W}} \gets \mathcal{V}[e,\,-|\mathcal{W}|:]$
    \State $\hat{\mathcal{I}} \gets \mathrm{Kara}(\mathcal{K}_{\mathcal{W}},\mathcal{V}_{\mathcal{W}}, r)$
    \State $(\mathcal{K}_{\mathcal{W}\setminus\mathcal{U}},\mathcal{V}_{\mathcal{W}\setminus\mathcal{U}})
    \gets
    (\mathcal{K}_{\mathcal{W}\setminus\mathcal{U}},\mathcal{V}_{\mathcal{W}\setminus\mathcal{U}})[\hat{\mathcal{I}}]$
    \State \textsc{MemoryCompaction}($e$)
    \State \textsc{UpdateMetadata}($e$)
\EndFor
    \EndIf
\EndWhile
\end{algorithmic}
\end{algorithm}

\section{Additional Experiment Results}
\subsection{Additional Results for Figure \ref{fig:F3}(a)}
\label{appx:f3}
\begin{figure}[ht]
  \centering
  \includegraphics[width=1.12\textwidth,trim = 40 40 -60 0]{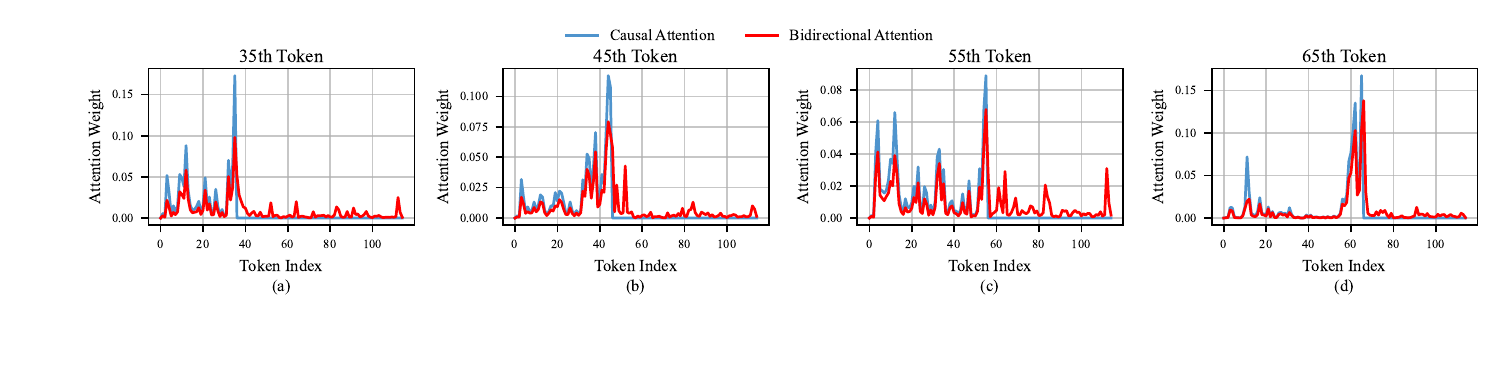}
  \includegraphics[width=1.12\textwidth,trim = 40 40 -60 0]{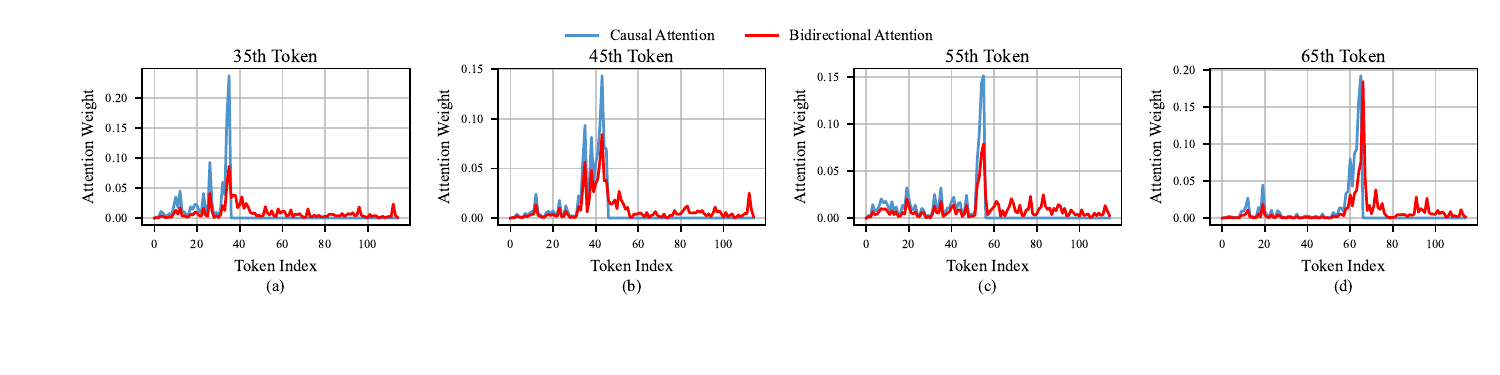}
  \caption{The top and bottom plots show the attention distributions of a token at Layer 10 and Layer 30, respectively.}
  \label{appfig:f3ab}
\end{figure}

We average the attention weights of each token’s multi-head KV pairs and visualize the results across different layers in Figure ~\ref{appfig:f3ab}.
We consistently observe token $x_i$’s causal attention distribution largely matches its bidirectional attention distribution on preceding positions.
\subsection{Visualization of Tokens Retained by Token2Chunk}
\label{appx:token2chunkres}
\begin{figure}[ht]
  \centering
  \includegraphics[width=1.\textwidth,trim = 0 0 -0 0]{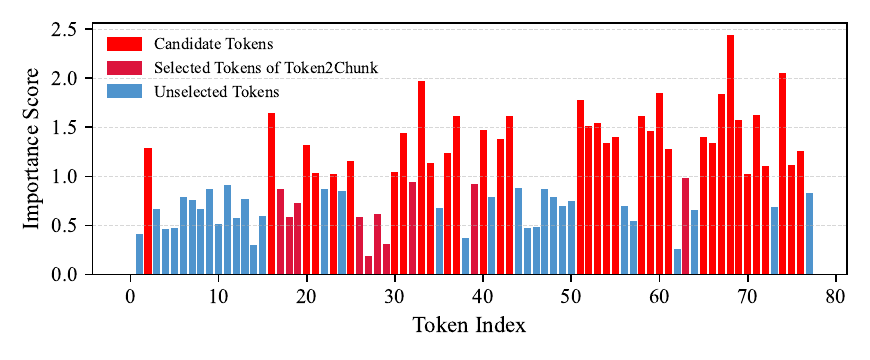}
  \caption{Visualization of the retained positions before and after applying Token2Chunk.
  We set the maximum chunk length and Token2Chunk budget to 7 and 10, respectively.}
  \label{appfig:f4ab}
\end{figure}
We analyze several sequences sampled from MATH-500 and report one representative example. 
We visualize the importance scores of the first 80 tokens and indicate which positions are additionally selected by Token2Chunk. 
In this example, bidirectional attention scoring selects 35 discrete tokens, and we set the maximum chunk length and Token2Chunk budget to 7 and 10, respectively.
We show the results for one representative layer and head. 
As shown in Figure \ref{appfig:f4ab}, Token2Chunk helps recover contiguous semantic information that may be missed by discrete KV-pair selection alone.
\subsection{Additional discussion of the results in Figure \ref{fig:sratio}}
\label{appx:dis_mainres}
We observe that RKV can be close to, or occasionally better than Kara in a few settings. 
We attribute this behavior to a difference in the effective \emph{uncompressed inference length} before compression is activated. 
Concretely, for a given prompt, Kara starts compression once the number of uncompressed tokens reaches $|\mathcal{W}|$, whereas RKV typically starts compression only when the uncompressed cache length reaches a predefined threshold. 
In common configurations, this threshold is substantially larger than $|\mathcal{W}|$ (\eg $2048$ vs \ $384$), which results in a longer uncompressed decoding trajectory for RKV. 
As a result, RKV may exhibit performance comparable to Kara in some cases.

However, RKV still relies on a threshold-triggered policy, which can incur frequent compression events under memory-constrained concurrent serving and degrade throughput. In contrast, Kara can be integrated into KvLLM with a periodic compression schedule. In this setting, we can increase $|\mathcal{W}|$ to extend the uncompressed trajectory while controlling when compression is triggered, improving the throughput--quality trade-off under concurrency.
\subsection{Results for NIAH Evaluation}
\label{appx:exp_niah}
\begin{figure*}[t]
    \centering    \includegraphics[width=1\linewidth, trim = 10 10 10 0]{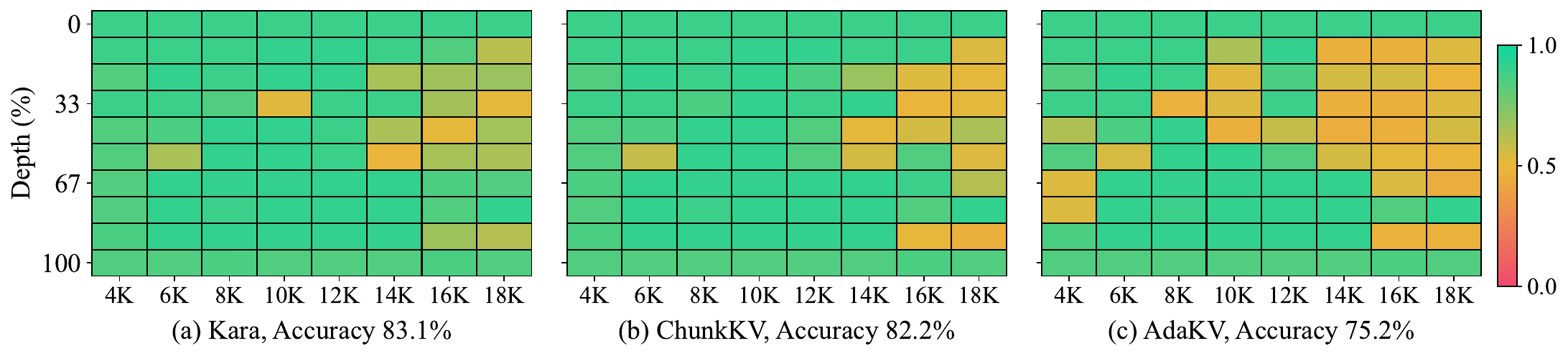}
    \caption{Needle-In-A-Haystack (NIAH) performance. 
    The x-axis denotes the input context length and the y-axis denotes the needle insertion depth. 
    Each cell reports the retrieval score for the corresponding (length, depth) setting, and we also report the mean accuracy averaged over all cells.}
    \label{fig:niah}
    \vspace{-10pt}
\end{figure*}
In this experiment, a short needle statement is inserted into a long context at varying depths, and the model is queried to retrieve the needle. 
We follow standard NIAH setup \cite{niah} with context lengths up to 18,000 and report accuracy averaged over 10 depths. 
Figure \ref{fig:niah} compares Kara with ChunkKV and AdaKV across different context lengths and insertion depths. 
\subsection{Latency and Throughput of KvLLM Framework}
\label{appx:kvllm}
We report throughput and overall latency under different batch sizes for KvLLM, vLLM+SnapKV, and vanilla vLLM in Table \ref{tab:batch_results}.
Note that when batch size is 16, 32, and 64, all frameworks are not memory-constrained, and all frameworks can maintain the maximum concurrency throughout decoding.
As shown in Table \ref{tab:batch_results}, we observe that 1) at batch size 16, vLLM+SnapKV achieves the best throughput. 
This is because compression is not triggered frequently under small batch size, and compression reduces KV cache memory usage and thus reduces memory access latency, leading to higher throughput. 
In contrast, KvLLM shows no clear improvement over vanilla vLLM in this setting, as KvLLM introduces additional recomputation overhead and the resource constraint is not binding. 2) At batch sizes 32 and 64, neither vLLM+SnapKV nor KvLLM provides throughput gains. 
One possible reason is that the increased batch size raises the compression overhead itself (\eg query cache storage in SnapKV and recomputation in KvLLM), offsetting potential benefits.
3) At batch sizes 128 and 256, KvLLM yields clear throughput improvements. 
In these memory-constrained settings, the available KV cache memory decreases rapidly during decoding, causing many sequences to wait and reducing realized concurrency. KvLLM releases PagedAttention blocks through periodic compression, allowing more sequences to participate in decoding and improving throughput. 
We can also tune the compression period to ensure that the compression overhead is covered by the resulting gains.

\begin{figure}[t]
    \centering
\setlength{\abovecaptionskip}{-10pt}    \includegraphics[width=0.45\textwidth]{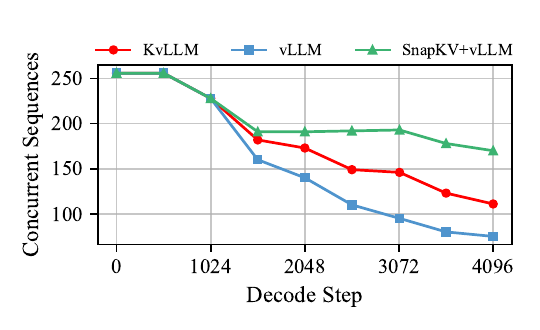}
    \caption{Comparison of concurrent sequence numbers for different frameworks.}
    \label{appxfig:concurrent}
\end{figure}

We also record the number of active running sequences across different decoding steps at batch size 256 in Figure \ref{appxfig:concurrent}.
KvLLM maintains substantially higher active concurrency than vanilla vLLM at batch size 256, which is consistent with the observed throughput improvements.
\begin{table}[ht]
\centering
\caption{The overall latency (Lat.) in seconds and average throughput (Thr.) of different frameworks under various batch sizes (\ie predefined maximum running sequences).}
\label{tab:batch_results}
\renewcommand{\arraystretch}{1.1}
\setlength{\tabcolsep}{4pt}
\small
\resizebox{\textwidth}{!}{
\begin{tabular}{l|cc|cc|cc|cc|cc}
\toprule
\textbf{Batch Size}
  & \multicolumn{2}{c|}{16}
  & \multicolumn{2}{c|}{32}
  & \multicolumn{2}{c|}{64}
  & \multicolumn{2}{c|}{128}
  & \multicolumn{2}{c}{256} \\ 
\midrule
\textbf{Metric}
  & Lat. (s) & Thr. (tokens/s)
  & Lat. (s) & Thr. (tokens/s)
  & Lat. (s) & Thr. (tokens/s)
  & Lat. (s) & Thr. (tokens/s)
  & Lat. (s) & Thr. (tokens/s) \\
\midrule
vLLM
  & {2,263} & {390}
  & \textbf{1,558} & \textbf{707}
  & \textbf{1,155} & \textbf{1,457}
  & 884  & 2,633
  & 678  & 2,943 \\
vLLM + SnapKV
  & \textbf{2,255} & \textbf{403}
  & 1,562 & 704
  & 1,171 & 1,377
  & 920  & 2,341
  & 738  & 2,556 \\
KvLLM
  & 2,260 & 393
  & 1,569 & 701
  & {1,163} & {1,421}
  & \textbf{864}  & \textbf{2,912}
  & \textbf{640}  & \textbf{3,392} \\
\bottomrule
\end{tabular}
}
\end{table}
\subsection{Complexity Analysis of Kara}
\label{appx:complexity}
We analyze the computation complexity of applying Kara once to a single sequence, ignoring memory access and KV movement.
Let the current sequence length be $S$, the sliding-window length be $W=|\mathcal{W}|$, and the protected buffer length be $U=|\mathcal{U}|$.
Thus, the computational complexity of Kara can be formulated as:
\begin{equation}
T_{\mathrm{Kara}}
=
O\!\left(
LWSD + LWD^2 + LW(W-U)D + LH(W-U)
\right),
\end{equation}
where the first term denotes the attention cost of the additional recomputation forward pass over the current window, the second term denotes the query projection and FFN cost in the recomputation, the third term denotes the cost of sliding-window bidirectional attention for KV importance estimation, and the fourth term denotes the complexity of Token2Chunk.

SnapKV has a computation complexity of
\begin{equation}
T_{\mathrm{SnapKV}}
=
O\!\left(
LU(S-U)D
\right),
\end{equation}
where $U$ denotes the query-buffer length.

\end{document}